# Securing Social Spaces: Harnessing Deep Learning to Eradicate Cyberbullying

Rohan Biswas[1[0009-0001-8840-2746]], Kasturi Ganguly[1[0009-0004-6976-5180]] ,

Arijit Das[1*[0000-0002-7202-5800]] , Diganta Saha[1*[]]

[1] Department of Computer Science and Engineering, Jadavpur University,
Kolkata, West Bengal, India
[*] Member, IEEE
lncs@springer.com

**Abstract.** In today's digital world, cyberbullying is a serious problem that can harm the mental and physical health of people who use social media. This paper explains just how serious cyberbullying is and how it really affects individuals exposed to it. It also stresses how important it is to find better ways to detect cyberbullying so that online spaces can be safer. Plus, it talks about how making more accurate tools to spot cyberbullying will be really helpful in the future. Our paper introduces a deep learning-based approach, primarily employing BERT and BiLSTM architectures, to effectively address cyberbullying. This approach is designed to analyse large volumes of posts and predict potential instances of cyberbullying in online spaces. Our results demonstrate the superiority of the hateBERT model, an extension of BERT focused on hate speech detection, among the five models, achieving an accuracy rate of 89.16%. This research is a significant contribution to "Computational Intelligence for Social Transformation," promising a safer and more inclusive digital landscape.

## 1 Introduction

Digital communication tools, such as social media, websites, email, text messages, or other online platforms provide individuals with the opportunity to openly and publicly share their thoughts and emotions with others. In spite of the considerable advantages offered by digital communication tools, their misuse leads to significant and adverse consequences. Harmful behaviors such as cyberbullying, cyber-aggression, hate speech, offensive language, and various forms of negativity have been growing more prevalent in these forms of communication. Cyberbullying[26] can cause emotional distress, low self-esteem, isolation, and even lead to more severe consequences such as self-harm and suicidal thoughts.

Due to the immediate requirement and the absence of suitable systems to manage the challenge of hate speech online, numerous researchers are driven to automate the process of identifying offensive content. The Natural Language Processing (NLP)[27] community plays a significant role in this. BERT, which stands for Bidirectional Encoder Representations from Transformers, is a groundbreaking natural language processing (NLP) model introduced by researchers at Google AI in 2018. HateBERT and RoBERTa are advanced variations of the original BERT model. HateBERT is specialized in detecting hate speech and offensive language. It is trained with data sourced from communities with banned content, enhancing its ability to identify harmful language patterns where the BERT model is also proficient in understanding language context but not mainly focused on hate speech detection. RoBERTa achieves better language representation through improved context understanding by using larger training datasets and more iterations during pretraining.

Cyberbully detection is challenging due to the complex, context-dependent nature of online interactions. The diverse forms of communication, evolving language, subtle indicators, and cultural nuances make it hard for automated systems to accurately identify instances of cyberbullying. Additionally, imbalanced data, user privacy concerns, and the need to balance precision with recall further complicate the task. The dynamic and ever-changing online landscape, coupled with the potential for malicious users to craft content that evades detection, amplifies the difficulty of developing effective and reliable cyberbully detection methods.

BERT models tackle these difficulties by leveraging their deep contextual understanding of language. BERT's bidirectional architecture captures intricate language nuances and context, helping to identify subtle indicators of cyberbullying. HateBERT specifically enhances its training with data from banned online communities, making it more adept at recognizing offensive language. These models' pretraining and fine-tuning processes enable them to learn patterns indicative of cyberbullying, thus addressing the complexity of the task and contributing to more accurate and efficient detection in diverse online environments.

Automated cyberbully detection functions like a sensor by continuously monitoring online content, detecting potentially harmful language, and alerting users or moderators when such content is identified. When the system detects content that meets predefined criteria for cyberbullying, it triggers alerts or actions, such as notifying the platform's moderators, flagging the content for review, or even issuing warnings to users. This "sensor-like" approach allows platforms to proactively address harmful behavior and create safer online environments.

The novelties of this work are – a) Proposing a generalized state-of-the-art methodology for detecting cyberbullying comments in any languages, b) The proposed method works for non-dictionary, code-mixed, colloquial text containing misspelled and incomplete slang words as well which is generally very difficult to detect with classical NLP technique, c) The proposed system has achieved almost 90% accuracy in best case which is better than the benchmark system[9] run over same standard dataset. The proposed model can be very useful for intelligent crowdsensing in social media for detecting cyberbullying.

## 2 Literature Review

Researchers have proposed various techniques for cyberbullying detection across different languages and platforms. Ahmed et al. [1] introduced a hybrid neural network for Bengali, achieving 87.91% binary and 85% multiclass accuracy. Ranasinghe et al. [2] used cross-lingual embeddings for offense detection in multiple languages. Islam et al. [3] improved Bengali sentiment analysis with multi-lingual BERT. Balakrishnan et al. [4] explored Twitter psychological features for effective detection. Talpur et al. [5] used PMI-based methodologies for identifying cyberbullying occurrences on the Twitter platform. In another work by Muneer et al. [6] achieved 90.57% accuracy with Logistic Regression. Islam et al. [7] combined NLP and ML for abusive message detection. Samghabadi et al. [8] used BERT for aggression identification. Iwendi et al. [9] found BLSTM effective for insult detection. Traditional methods were outperformed by deep learning for cyberbullying in social networks [10]. Research [11] reviewed prediction models and ML techniques for cyberbullying detection. Hani et al. [12] achieved 92.8% accuracy with NN. Emon et al. [13] improved Bengali sentiment analysis using RNN. Chakraborty et al. [14] used SVM for Bengali abusive language detection. Tarwani et al. [15] achieved 80.26% accuracy for cyberbullying in Hinglish. Murnion et al. [16] collected in-game chat data to address cyberbullying. Al-Ajlan et al. [17] introduced CNN-CB with 95% accuracy. Research [18] addressed Arabic cyberbullying detection. Zhang et al. [19] proposed PCNN for detecting misspelled cyberbullying. Al-Garadi et al. [20] achieved high accuracy in Twitter cyberbullying detection using unique features.

## 3 Methodologies

In this section, we present the implementation details of our experimental setup. Here 3 transformer models and 1 deep learning model are used. the implemented approach that is shown in Fig 1.

At first two datasets are merged to get the desired dataset. Then the dataset is preprocessed and produces tokenized data. That tokenized data is splitted into training, validation and test sets. Word embeddings are generated for each set. Then the model is trained and it generates accuracy and evaluation metrics using the test set.

Here is a brief description about the models which are used for the experiment.

### 3.1 BERT

Built upon the transformer architecture, Bidirectional Encoder Representation from Transformers (BERT) [1] leverages an extensive corpus of unlabeled text, including Wikipedia (2500 million words) and Book Corpus (800 million words), for its pre-training. The significant achievement of BERT predominantly arises from its pre-training phase, during which it learns from an extensive collection of texts. The BERT model assimilates information from both the preceding and subsequent parts of a sentence context.

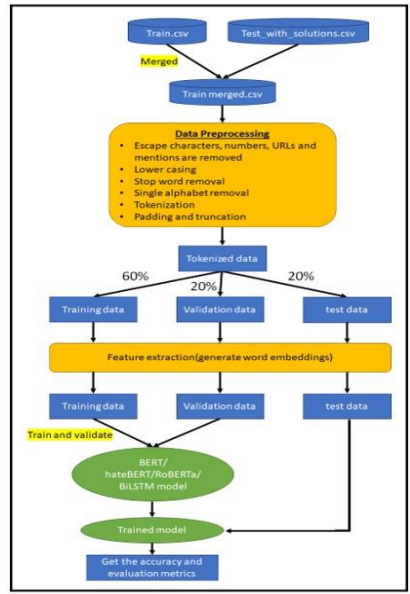

**Fig. 1.** Implemented approach

### 3.2 hateBERT

HateBERT[22] is a model built upon the foundation of the BERT architecture, which is specifically designed to address issues related to hate speech detection and offensive language identification in text. HateBERT is an enhanced version of the English BERT base uncased model. It's created by extending the training of the original model using over a million posts sourced from Reddit communities that have been banned due to their offensive content. By leveraging the inherent contextual understanding capabilities of BERT, HateBERT aims to accurately identify and classify instances of hate speech, offensive content, and potentially harmful language within text data.

### 3.3 RoBERTa

RoBERTa[23], short for "A Robustly Optimized BERT Pretraining Approach," is a derivative of BERT model that retains the same architecture as BERT but introduces some modifications in the training procedure. It employs a larger batch size and trains on more data for a longer duration. This extended training duration helps RoBERTa to thoroughly capture context and nuances in the text. Unlike BERT, which uses masked language modeling as one of its pretraining tasks, RoBERTa trains solely using a variant of the masked language modeling task.

### 3.4 BiLSTM

Bidirectional Long Short-Term Memory (BiLSTM)[25] is designed to capture sequential dependencies in data, particularly in sequences of variable length, such as sentences or time series data. BiLSTM networks offer a valuable tool for processing sequential data, especially in NLP tasks where understanding the context from both directions is crucial for accurate predictions.

## 4 Experiments and Results

The dataset[21] contains labeled comments with two attributes: the time the comment was made and the content of the comment in English language, sometimes with formatting. The dataset consists of two files, "train.csv" and "test_with_solutions.csv." The "train.csv" file contains a total of 3947 tweets, with class labels 0 or 1 indicating Neutral or Insulting, respectively. It includes 1049 Insulting tweets and 2898 Neutral tweets. The "test_with_solutions.csv" file contains 2647 tweets, comprising 693 Insulting tweets and 1954 Neutral tweets. These two files are combined into one file named "train_merged.csv".

This dataset is split into train, test, and validation sets using a 60:20:20 ratio to assess the model's performance comprehensively.

The results of all different experiments that we performed are presented in Table 1.

In this research, hateBERT outperforms other 4 models considering evaluation metrics for detection of cyberbullying.

For hateBERT model,
   the neutral class prediction result is as follows,
      Precision = 0.94
      Recall = 0.91
      F1-Measure = 0.93.
   the Insult class result prediction is as follows,
      Precision = 0.76,
      Recall = 0.83
      F1-Measure = 0.79
For RoBERTa model,
   the neutral class prediction result is as follows,
      Precision = 0.86
      Recall = 0.96
      F1-Measure = 0.91.
   the Insult class result prediction is as follows,
      Precision = 0.84,
      Recall = 0.57
      F1-Measure = 0.68

The hatebert model shows the accuracy of 89.16% where the RoBERTa model shows 85.59% accuracy.

Clearly RoBERTa shows the second best accuracy among these models. For the BiLSTM model, two different cases i.e. without any embedding and with FastText embedding are shown and from the accuracy it is clear that embedded BiLSTM works better.

Table 1 presents the precision, recall and F1-measure values for neutral and insulting classes and macro average F1 and accuracy for all 4 models and baseline paper[9]'s result is also defined in the table.

**Table 1.** Classification report and accuracy for all models

| Model | Labels | Precision | Recall | F1 Score | Macro Average F1 Score | Accuracy |
| --- | --- | --- | --- | --- | --- | --- |
| BERT | Neutral | 0.88 | 0.91 | 0.89 | 0.78 | 83.78% |
| | Insulting | 0.72 | 0.65 | 0.68 | | |
| HateBERT | Neutral | 0.94 | 0.91 | 0.93 | 0.86 | 89.16% |
| | Insulting | 0.76 | 0.83 | 0.79 | | |
| RoBERTa | Neutral | 0.86 | 0.96 | 0.91 | 0.80 | 85.59% |
| | Insulting | 0.84 | 0.57 | 0.68 | | |
| BiLSTM(without Embedding) | Neutral | 0.84 | 0.94 | 0.89 | 0.76 | 82.49% |
| | Insulting | 0.76 | 0.53 | 0.63 | | |
| BiLSTM(with FastText Embedding) | Neutral | 0.87 | 0.90 | 0.88 | 0.78 | 83.32% |
| | Insulting | 0.72 | 0.65 | 0.68 | | |
| Baseline Paper[9] Results using BiLSTM | Neutral | 0.86 | 0.91 | 0.88 | 0.76 | 82.18% |
| | Insulting | 0.71 | 0.60 | 0.65 | | |

For better understanding, Figure 2 shows the comparison of Precision, Recall and F1-Measure for all 4 models.

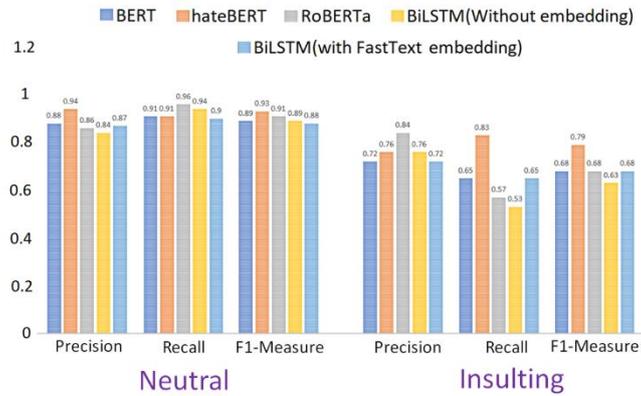

**Fig. 2.** : Classification report

Figure 3 shows the comparative discussion of 4 different models' accuracy and Figure 4 shows the comparison between our best result and baseline paper[9] result .

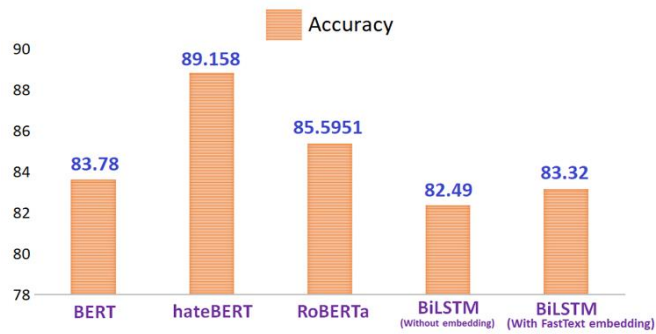

**Fig. 3.** : Classifiers accuracy

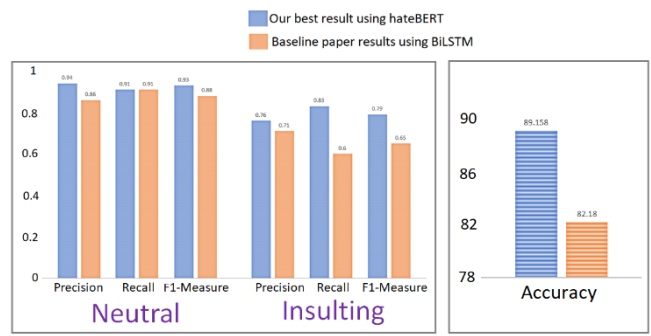

**Fig. 4.** : Comparison with Baseline paper[9]

## 5      Conclusion

Cyberbullying can have serious negative impacts on the victims' mental and emotional well-being. In some cases, cyberbullying has even resulted in tragic consequences, such as self-harm or suicide. Detecting and addressing cyberbullying helps protect individuals from harm and reduces the negative impact on their mental well-being. Various approaches and techniques that can be used for cyberbullying detection including machine learning and NLP. Numerous research studies have been conducted on cyberbullying detection using Natural Language Processing (NLP) techniques[24].

In this study, three transformer models, namely BERT, hateBERT, RoBERTa and BiLSTM models are used for experiments. Data preprocessing steps that are applied include : text cleaning, stop word and single alphabet removal, tokenization, Padding and truncation. hateBERT model achieved the highest accuracy and F1-Measure compared to other models. hateBERT model showed 89.158% accuracy where BERT, RoBERTa and BiLSTM could achieve 83.78%, 85.5951% and 83.32% accuracy respectively.

In the future, we shall explore advanced techniques like oversampling, undersampling, data augmentation, Class Weighting, Ensemble Methods etc. to handle the class imbalance in the dataset and improve the model's performance on the minority class. We can use hybrid models like combination of BERT's contextual embeddings with Bidirectional LSTM (BLSTM). Hyperparameters can be optimized. We can also develop a real-time implementation of the model for dynamic monitoring and immediate response to cyberbullying instances, providing a proactive approach to online safety.

## Acknowledgement

We extend our appreciation to our supervisors and mentors Dr. Arijit Das and Professor Dr. Diganta Saha for their invaluable guidance. We are deeply grateful to my friends Rupam Saha and Imanur Rahaman for their invaluable assistance in the research work. We would also like to acknowledge the support of Jadavpur University and our families unwavering support. We would like to thank Gautam Srivastava  for sharing the dataset[21].